%% file: landShare.tex
\documentclass[dvipsnames,format=sigconf,anonymous=false,review=false]{acmart} 

\def\vec#1{\mathchoice{\mbox{\boldmath$\displaystyle#1$}}
{\mbox{\boldmath$\textstyle#1$}}
{\mbox{\boldmath$\scriptstyle#1$}}
{\mbox{\boldmath$\scriptscriptstyle#1$}}}

\usepackage{booktabs} 
\usepackage{amsmath}
\usepackage{algorithm}
\usepackage[noend]{algpseudocode}
\usepackage[T1]{fontenc}
\usepackage{tabularx}
\usepackage{multirow}
\usepackage{array}
\usepackage{pbox}

\usepackage{color,soul}
\usepackage{nameref}

\usepackage{pgfplots}
\usepgfplotslibrary{statistics}
\usepackage{graphicx}
\usepackage{subcaption}
\usepackage{tikz}

\usepackage{url}
\usepackage{bm}

\definecolor{darkgreen}{rgb}{0.30, 0.50, 0.0}

\makeatletter
\renewcommand{\ALG@name}{Pseudocode}
\makeatother

\newcolumntype{L}[1]{>{\raggedright\let\newline\\\arraybackslash\hspace{0pt}}m{#1}}
\newcolumntype{C}[1]{>{\centering\let\newline\\\arraybackslash\hspace{0pt}}m{#1}}
\newcolumntype{R}[1]{>{\raggedleft\let\newline\\\arraybackslash\hspace{0pt}}m{#1}}

\graphicspath{{images/}}

\setcopyright{acmcopyright}
\acmDOI{10.1145/nnnnnnn.nnnnnnn} 
\acmISBN{978-x-xxxx-xxxx-x/YY/MM} 
\acmConference[GECCO '25]{The Genetic and Evolutionary Computation Conference 2025}{July 14--18, 2025}{Málaga, Spain}
\acmYear{2025}

\copyrightyear{2025}
\acmYear{2025}
\setcopyright{cc}
\setcctype{by}
\acmConference[GECCO '25]{Genetic and Evolutionary Computation
Conference}{July 14--18, 2025}{Malaga, Spain}
\acmBooktitle{Genetic and Evolutionary Computation Conference (GECCO '25),
July 14--18, 2025, Malaga, Spain}\acmDOI{10.1145/3712256.3726419}
\acmISBN{979-8-4007-1465-8/2025/07}

\hypersetup{draft}
\begin{document}
	\title{Seeking and leveraging alternative variable dependency concepts in gray-box-elusive bimodal land-use allocation problems}

    \author{Jakub Maciążek}
 	\affiliation{
 		\institution{Dep. of Systems and Comp. Networks}
 		\institution{Wroclaw Univ. of Science and Techn.}
 		\city{Wroclaw}
 		\country{Poland} 
 	}
 	\email{kuba.maciazek@gmail.com}
	
 	\author{Michal W. Przewozniczek}
 	\affiliation{
 		\institution{Dep. of Systems and Comp. Networks}
 		\institution{Wroclaw Univ. of Science and Techn.}
 		\city{Wroclaw}
 		\country{Poland} 
 	}
 	\email{michal.przewozniczek@pwr.edu.pl}

        \author{Jonas Schwaab}
 	\affiliation{
 		\institution{Inst. for Spatial and Landscape Plann.}
 		\institution{ETH Zürich}
 		\city{Zürich}
 		\country{Switzerland} 
 	}
 	\email{jonasschwaab@ethz.ch}

	\renewcommand{\shortauthors}{}

	\begin{abstract}

        Solving land-use allocation problems can help us to deal with some of the most urgent global environmental issues. Since these problems are NP-hard, effective optimizers are needed to handle them. The knowledge about variable dependencies allows for proposing such tools. However, in this work, we consider a real-world multi-objective problem for which standard variable dependency discovery techniques are inapplicable. Therefore, using linkage-based variation operators is unreachable. To address this issue, we propose a definition of problem-dedicated variable dependency. On this base, we propose obtaining masks of dependent variables. Using them, we construct three novel crossover operators. The results concerning real-world test cases show that introducing our propositions into two well-known optimizers (NSGA-II, MOEA/D) dedicated to multi-objective optimization significantly improves their effectiveness.
        \end{abstract}
	
	%
	%


	
	\maketitle

\input{landShareBody}

	\bibliographystyle{ACM-Reference-Format}
	\bibliography{landShare} 
	
\end{document}

%% file: landShareBody.tex
\section{Introduction}
\label{sec:intro}

Land-use optimization (LUO) is used to address some of the most pressing environmental challenges of the 21\textsuperscript{st} century, including climate change, the biodiversity crises and the loss of fertile agricultural land \cite{envirStrass}. LUO problems consist in determining the optimal locations for different land-use types such as urban, agriculture and forest. This process usually entails balancing conflicting objectives. For example, the revenue from urban development has to be traded-off against potential losses of biodiversity and agricultural soils. As a consequence, multi-objective optimization approaches have become a standard approach for solving land-use allocation problems \cite{envirRahman}. The objectives in multi-objective LUO can be broadly categorized into two groups. The first group comprises objectives that evaluate the suitability of a particular land-use type at a given location based on the location’s characteristics. The second group includes objectives that assess suitability based on the surrounding land-use types and the overall spatial pattern of land-use. The latter are the most common objectives in LUO \cite{envirRahman}, but due to their complexity, they pose a major challenge to solving LUOs. In this study, we show that previous attempts to deal with these challenges \cite{envirSchwaab}, can be substantially improved to solve LUOs more efficiently. 

To propose significantly more effective optimizer than the currently available, we refer to the idea of gray-box optimization. Gray-box optimization aims at proposing highly effective optimizers by using variable dependencies to leverage variation operators \cite{GrayBoxWhitley}. For binary-encoded problems, it is frequent for gray-box-oriented studies to use Walsh decomposition to obtain such dependencies \cite{walshSurBin}. Thus, gray-box optimizers use generalized problem-specific knowledge to gain effectiveness, yet they remain general enough to remain applicable to many different problems \cite{whitleyNext}. \par

In this work, we adopt solution encoding and decoding mechanisms from the state-of-the-art literature \cite{luoCoreSchwaab}, which results in using binary-encoded solutions. Some solutions may be infeasible and must be repaired. However, problem features make deterministic repair mechanisms poorly promising, and random procedures are employed for this purpose. Therefore, we can not deterministically evaluate infeasible solutions by evaluating their repaired version while leaving the original (infeasible) solution unchanged. Due to this reason, using Walsh decomposition is unavailable because it requires evaluating infeasible solutions. Thus, despite being binary encoded, the considered problem can not be decomposed by Walsh decomposition and can be considered as \textit{gray-box-elusive}.\par

Therefore, we propose problem-dedicated variable dependency concepts to utilize the potential brought by dependency-aware variation operators despite LUO being a grey-box-elusive problem. Using the proposed variable dependency concepts, we propose a problem-dedicated crossover and several other mechanisms. The experiments confirm that introducing our propositions into two optimizers dedicated to multi-objective (MO) optimization significantly improves their effectiveness.\par

The rest of this work is organized as follows. The next section presents the background of our research. Section \ref{sec:problem} presents the definition of the considered problem, solution encoding, and standard operators. In Sections \ref{sec:operators}-\ref{sec:inits}, we describe in detail the proposed operators and explain their core intuitions. The sixth section presents the experiment results. Finally, the last section concludes this work and points out the main directions for future work.

\section{Related Work}
\label{sec:relWork}

\subsection{Land Use Allocation}
\label{sec:rw:LUO}

LUO has been used very frequently to optimize the location of nature conservation and restoration areas, of urban development areas and of areas for renewable energies \cite{envirStrass,envirRahman}. LUO is used to provide decision-support for spatial planners and policy-makers operating at international, national, regional and local level \cite{envirSchwaab}. Land use and land-use changes have an impact on societal, economic and environmental objectives \cite{envirChikumbo}. Thus, it is in most cases mandatory to consider them as multi-objective optimization problems. \par

A wide variety of approaches have been adopted to formulate multi-objective LUO problems and to solve them. In most cases, the spatial units to be optimized are represented in a regular grid (i.e. a pixel-based map), but there are also examples of relying on vector based (i.e. spatial polygon) representations \cite{envirStewart}. The decision-variables are usually discrete representations of different land-uses (i.e. different integer values for each land-use type) and are in many cases a binary variable representing the choice between two land-uses at a specific location \cite{envirRahman}. 

The most common constraints are that a specific land-use should meet or not exceed a predefined area (Schwaab et al., 2017). There are a wide range of different objectives included that can broadly been categorized as either being additive or spatially dependent (Stewart and Janssen, 2014). Spatially dependent objectives include measures of contiguity (compactness, connectedness, etc.), distances between different land-use types and a large variety of landscape metrics measuring, e.g., fragmentation. \par

To solve LUOs, many different approaches have been employed, including Pareto- and decomposition-based approaches \cite{envirKaim}. Due to the complexity of most LUO problems and their multi-objective nature, the most common way of solving these problems has been to use evolutionary algorithms \cite{envirRahman}. Thus, the real-world LUO problem that we are using in this study includes some of the most common features of LUO problems, and hence, we believe that the presented advances can be useful to a wide range of LUO applications.

\subsection{Multi-objective optimization}
\label{sec:rw:mo}

Binary-encoded multi-objective (MO) problems require optimizing $m$ objective functions simultaneously $f(\vec{x}) = (f_1(\vec{x}),\ldots,f_{m}(x))$, where $\vec{x}=(x_1,x_2,\ldots,x_{n})$ is a binary vector of size $n$. Assuming that the optimized function is minimized, solution $\vec{x^0}$ \textit{dominates} $\vec{x^1}$ if and only if $f_i(\vec{x^0}) \leq f_i(\vec{x^1})$ $\forall{i}\in \{1,2,\ldots,m\}$ and $f(\vec{x^0})\neq f(\vec{x^1})$. Pareto-optimal set $\mathcal{P}_S$ is a set of all non-dominated solutions (Pareto-optimal solutions). The objective value vectors of all Pareto-optimal solutions form a Pareto-optimal front $\mathcal{P}_F$. Frequently, the size of $\mathcal{P}_F$ is large. Therefore, in MO, we wish to find a PF that approximates $\mathcal{P}_F$ well \cite{MoGomeaSwarm}.\par

Non-dominated Sorting Genetic Algorithm II (NSGA-II) \cite{nsga2} is a well-known MO-dedicated optimizer. NSGA-II uses a population of individuals. To compare their quality, the domination relation is used to cluster them into subpopulations. The first subpopulation (front) consists of non-dominated solutions in the clustered population. The second front is formed by the solutions that are non-dominated after removing the first front from the population, and so on. The higher the front number, the lower the quality of the solutions it groups. If compared solutions are a part of the same front, NSGA-II uses the \textit{crowding distance} measure. Solutions related to high values of crowding distance are considered original (in spite of the given population) and, therefore, valuable. The details concerning NSGA-II can be found in \cite{nsga2}.\par

In MOEA/D \cite{moead}, each solution is scalarized. Thus, each solution solves a different single-objective problem. This way, MOEA/D avoids using the domination relation, which may be an advantage while solving problems with many objectives \cite{pfQuality}. In each iteration, for each solution, a set of candidate solutions is generated, and the best of them (in terms of scalarized fitness) replaces the original solution if its scalarized fitness is higher. MOEA/D uses the \textit{mating restrictions} mechanisms \cite{matingRestr}, which limits the range of solutions with which a given solution can mix. Mating restrictions allow for avoiding mixing individuals that solve significantly different scalarized problems. Many state-of-the-art studies use MOEA/D or its improved versions as the research starting point \cite{moeadStrong,moeadStrong2,moeadStrongIGD}.

\subsection{Dependency-based optimization leveraging}
\label{sec:rw:gray}
In this work, we consider a gray-box-elusive problem, which can not be decomposed using typical gray-box concepts. However, the objective of our research was to enable the potential brought by dependency-using operators in the optimization of the considered problem. Therefore, we present the problem decomposition concepts in more detail.\par

Gray-box optimization focuses on proposing operators that use knowledge about variable dependencies to improve optimizers' effectiveness \cite{whitleyNext}. The concept of variable dependency can be interpreted in many ways, e.g., \cite{linc,dled,2dled}. However, in gray-box-related works, it is frequent to consider non-linear dependencies, i.e., variables $x_g$ and $x_h$ are considered dependent if there exists $\vec{x}$ such that \cite{linc}:
    \begin{equation}
    \small
    \label{eq:nonLinear}
        f(\vec{x}) + f(\vec{x}^{g,h}) \neq f(\vec{x}^g) + f(\vec{x}^h)
    \end{equation}
    where by $\vec{x}^{g}$, $\vec{x}^h$, and $\vec{x}^{g,h}$, are solutions obtained from $\vec{x}$ by flipping gene $g$, gene $h$ or both of them, respectively.\par

Non-linear dependencies can be also obtained using \textit{Walsh decomposition} \cite{heckendorn2002} that allows defining any pseudo-Boolean function as:

    \begin{equation}
    \small
    \label{eq:walsh-decomposition}
    f(\vec{x}) = \sum_{i=0}^{2^n-1} w_i \varphi_i(\vec{x}) 
    \end{equation}
    where $w_i \in \mathbb{R}$ is the $i$th Walsh coefficient, $\varphi_i(\mathbf{x}) =(-1)^{\mathbf{i}^\mathrm{T}\mathbf{x}}$ generates a sign, $\mathbf{i} \in \{0,1\}^n$ is the binary representation of index $i$, and $\vec{x}=\{x_1,x_2,\ldots,x_{n}\}$ is a binary solution vector of size \textit{n}.  \par

In spite of Walsh decomposition, two variables $x_g$ and $x_h$ are dependent if there exists at least one nonzero Walsh coefficient $w_i$ such that the $g$th and $h$th elements of $\mathbf{i}$ are equal to one \cite{ilsDLED}. The dependencies identified by Walsh decomposition are equivalent to non-linearity check.\par

The obtained dependencies are usually stored in a Variable Interaction Graph (VIG), which is a square matrix. Each of its entries equals one if the variable pair it refers to is dependent or zero otherwise. Gray-box operators cluster dependencies stored in VIG to obtain masks of dependent variables that are used in variation operators, e.g., as a mixing mask in Partition Crossover (PX) \cite{pxForBinary} or as a perturbation mask in iterated local search \cite{ilsDLED}.\par

In black-box optimization, the \textit{a priori} knowledge about variable dependencies is unavailable. Therefore, it must be discovered during the optimizer run \cite{ltga}. The recent propositions of linkage learning techniques allow for effective and efficient dependency discovery \cite{FIHCwLL,ilsDLED} that enables using gray-box operators in black-box optimization \cite{dgga}. Some studies, focus on obtaining the surrogate of the optimized function to enable no-cost or low-cost fitness evaluations and a VIG at the same time \cite{ellGomea,walshSurBin,walshSurBin2}.\par

For some problems, the non-linearity check may be oversensitive. Therefore, the non-monotonicity check was proposed in \cite{GoldMonotonicity}, which finds $x_g$ and $x_h$ dependent if one of the conditions is \textbf{not} true:

    \begin{equation}
        \small
        \label{eq:nonMonoGold1}
        \begin{aligned}
            \textbf{if}(f(\vec{x^g}) > f(\vec{x})) & \textbf{and} (f(\vec{x^h}) > f(\vec{x})) \textbf{then} \\ 
            & (f(\vec{x^{g,h}}) > f(\vec{x^g})) \textbf{and} (f(\vec{x^{g,h}}) > f(\vec{x^h}))
        \end{aligned}
    \end{equation}

    \begin{equation}
        \small
        \label{eq:nonMonoGold2}
        \begin{aligned}
            \textbf{if}(f(\vec{x^g}) < f(\vec{x})) & \textbf{and} (f(\vec{x^h}) < f(\vec{x})) \textbf{then} \\ 
            & (f(\vec{x^{g,h}}) < f(\vec{x^g})) \textbf{and} (f(\vec{x^{g,h}}) < f(\vec{x^h}))
        \end{aligned}
    \end{equation}
 
 Using the non-monotonicity check allows to optimize effectively non-additively decomposable problems \cite{irrg}.

\section{Land Use Optimization problem}
\label{sec:problem}

\subsection{Problem definition and quality measures}
\label{sec:problem:def}

Here, we present the Land Use Optimization (LUO) problem in detail. We adopt notation similar to the one proposed in \cite{luoNotation}. In LUO, we consider the $RxC$ matrix of areas. Each area can fall into one of $K$ land-use categories. In the considered LUO instances, we are to propose a land-use map with $U$ urban areas by transforming $T$ agricultural areas into urban ones (the initial number of urban areas is $U-T$). Therefore, we wish to minimize the \textit{Loss of Agricultural Productivity} (LAP) \cite{problemTelSchwaab} caused by the transformation.

\begin{equation}	\label{measLap}
    LAP(\vec{y})= 1/A\sum_{r=1}^R \sum_{c=1}^C \sum_{k=1}^K a_{rck}y_{rck}
\end{equation}
subject to
\begin{equation}	\label{measLapControl}
    \forall_{r=1..R; c=1..C} \sum_{k=1}^K y_{rck} = 1
\end{equation}
\begin{equation}	\label{measLapUrbanSum}
    \sum_{r=1}^R \sum_{c=1}^C y_{rc1} = U
\end{equation}
where $\vec{y}$ denotes the considered land-use map expressed by binary variables $y_{rck}$, $U$ is the number of desired urban areas, $k=1$ is the urban area type, $a_{rck}$ is the potential loss of agricultural productivity for a given area, and $A$ is the sum of potential loss of agricultural productivity for whole matrix. \par

Except minimizing the loss of agricultural productivity, we wish to keep all types of areas compact. Therefore, we wish to minimize the \textit{Total Edge Length (TEL)} \cite{problemTelSchwaab} measure.


\begin{equation}	\label{measLap}
    TEL(\vec{y})= \sum_{r=1}^R \sum_{c=1}^C \sum_{k=1}^K (y_{r+1,c,k}+y_{r-1,c,k} + y_{r,c+1,k} + y_{r,c-1,k})\cdot y_{rck} 
\end{equation}
subject to
\begin{equation}	\label{measLapCOnstr}
    \forall_{k=1,...,K;r \in \{0, R+1\};c \in \{0,C+1\}} y_{rck} = 0
\end{equation}

LUO is classified as NP-hard problem \cite{LUOnpHard,LUOnpHard2}.

\subsection{Solution encoding and repair mechanisms}
\label{sec:problem:encoding}
In LUO, we can represent each solution as the binary vector $f(\vec{x})$, where $\vec{x}=\{x_1,x_2,\ldots,x_{n}\}$ of size $n$. Each $x_i$, we interpret as a decision "do not transform associated agricultural area" ($x_i=0$) or "transform associated agricultural area into urban" ($x_i=1$). Note that $n$ equals the initial number of agricultural areas. If the solution is feasible, then $u(\vec{x}) = T$, where $u(\vec{x})$ is the \textit{unitation} of $\vec{x}$ \cite{decFunc} (the number of ones).

\begin{figure}[h]
    \centering
    \includegraphics[width=0.95 \linewidth]{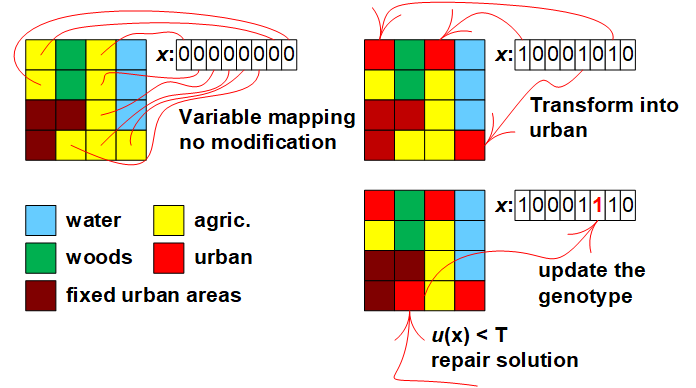}
    \caption{The example of variable mapping, solution encoding and repair for LUO with $R=C=4$, $K=5$, $T=4$, and $U=7$.}
    \label{fig:encodingAndRepair}
\end{figure}

In Fig. \ref{fig:encodingAndRepair}, we present the solution to the exemplary LUO instance. Each agricultural area is referred by each $x_i$. If $x_i=1$, then it modifies the original map by turning the area referred by $x_i$ from agricultural into urban area. If $u(\vec{x}) \neq T$, then the solution must be repaired. In Fig. \ref{fig:encodingAndRepair}, $u(\vec{x}) < T$. Therefore, an additional area must be transformed into urban. If $u(\vec{x}) > T$, then the repair mechanism would resign from some of the modifications defined by $\vec{x}$.\par

In \cite{luoCoreSchwaab}, two repair mechanisms are considered. Both are a part of mutation. In \textbf{Random Repair Mutation (RRM)}, we randomly select any $x_i=1$ and assign it with $0$ until $u(\vec{x}) = T$. Using \textbf{Biased Repair Mutation (BRM)} we can fix the deficiency or excess of urban areas. The areas to repair (to transform or to resign from transforming into urban areas) are chosen randomly, but the probability of choosing a given area is proportional to the number of neighbouring (von Neumann neighbourhood) areas of the cell post-repair category.

\subsection{Standard operators}
\label{sec:problem:StandOperators}
As a baseline we consider the operators proposed in \cite{luoCoreSchwaab}, where they were introduced into NSGA-II. In this work, we also consider these standard operators introduced into MOEA/D.\par

\textbf{Random Population Initialization (SP-I)} is a random initialization, which from the initial representation of given case problem, with random uniform distribution chooses \textit{T} agricultural areas to transform them into urban.\par

\textbf{Angle Crossover (AC)} operator generates offspring by combining halves of parents matrixes obtained via split line going through the centre of them at randomly varying angle.\par

\textbf{Combined Mutation (MutC))} sequentially uses five atomic operators. First, the \textbf{Random Block Mutation (RBM)} is executed with 10\% probability. RBM transforms a random size block of areas in random localization from agricultural into urban. It is followed by RRM, which ocures only if RBM was previously triggered. Next, again with 10\% probability \textbf{Random Cell Mutation (RCM)} transforms one randomly chosen urban area into agricultural and one randomly chosen rural agricultural area into urban one. Then, BRM is executed with 100\% probability. Finally, \textbf{Biased Cells Patch Mutation (BCPM)} is executed with 10\% probability. It transforms a continuous patch of adjacent urban areas into rural ones with a probability inversely proportional to its size. Next, with a probability proportional to its size, new urban patch of continuous areas is chosen. Number of previously removed urban areas is then added to it. Probability of conversion of an agricultural area to this patch depends on the number of adjacent cells of opposite category (urban ones), like in \textit{BRM}. MutC is guaranteed to output a feasible solution.\par


In this work, we also consider a modified MutC version. In \textbf{MutC2}, we skip RRM after RBM. Thus, the changes made by RBM will not be fixed by RRM but by BRM which is biased towards proposing better-fitting solutions solutions of better compactness.

\section{Proposed Crossover Operators}
\label{sec:operators}

Gray-box optimizers leverage their effectiveness using variables dependencies. However, both dependency checks described in Section \ref{sec:relWork}, namely non-linearity and non-monotonicity check, do not apply to the considered solution encoding because of the following reasons. In the considered LUO, we are to choose T areas to modify. Thus, if solution $\vec{x_a}$ is feasible, then solutions $\vec{x_a^g}$, $\vec{x_a^h}$, and $\vec{x_a^{g,h}}$ will be infeasible because their unitation is different to $u(\vec{x_a})$. At the same time, the repair procedures follow the Lamarck effect \cite{MuPPetsBaldwinEon}, i.e., the repair procedure modifies the original solution. Thus, in the considered setting that is typical for LUO \cite{envirRahman}, we can not compute $f(\vec{x_a^g})$, $f(\vec{x_a^h})$, and $f(\vec{x_a^{g,h}})$, which makes non-linearity and non-monotonicity checks unavailable. Using Walsh decomposition is unavailable due to the same reasons - many genotypes are infeasible and can not be evaluated. \par

Gray-box operators can lead to excellent results \cite{GrayBoxWhitley,ilsDLED} while preserving their simplicity and, frequently, being parameter-less \cite{pxForBinary}. Oppositely, the standard operators proposed in Section \ref{sec:problem:StandOperators}, seem sophisticated and require additional settings (e.g., some of suboperators in MutC and MutC2 are exectuted with a given probability).\par

Therefore, we propose a problem-dedicated concept of defining dependencies between variables. The core intuition behind our proposal is that in LUO we shall rather consider and process the regions of neighbouring areas of the same type, rather than single areas. Thus, for a given pair of mixed individuals, we cluster neighbouring areas considering three ways of filtering.\par

The core intuition behind our proposal is based on the observation that TEL is a spacial measure, i.e., a single variable (related to a single area) influences TEL not by its own value separately but rather by the relationship between its value and the values of variables that refer to the neighbouring areas. Therefore, we wish to create clusters of such variables and exchange them together. Such an intuition is the same to the assumptions standing by typical gray-box operators, e.g., PX \cite{pxForBinary,dgga}.\par

\begin{figure}[h]
    \centering
    \includegraphics[width=0.95 \linewidth]{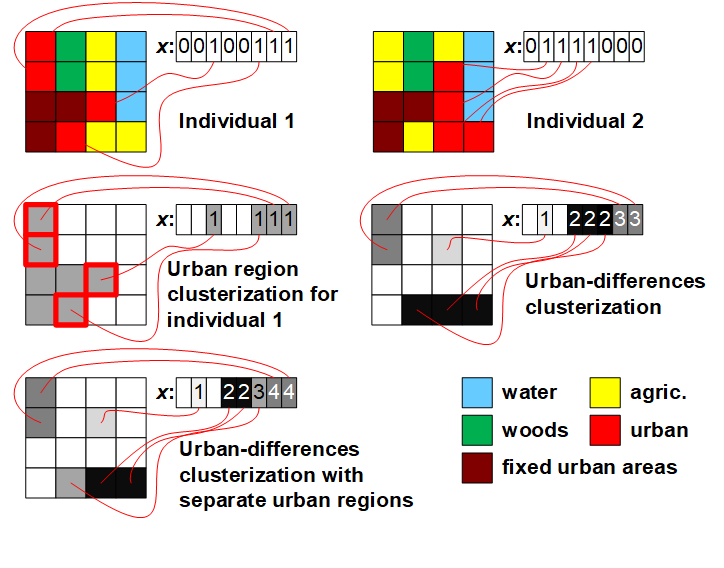}
    \caption{Clusterization examples employed by proposed crossovers (we consider the same instance as in Fig \ref{fig:encodingAndRepair})}
    \label{fig:depAndCross}
\end{figure}

In all of the proposed crossovers, we obtain clusters of dependent variables. During offspring creation, we process all clusters in a random order and assign the values from one or the other parent. This process resembles the uniform crossover, but we consider gene clusters instead of single genes.\par

Fig. \ref{fig:depAndCross} presents three variable clusterization ideas for crossover. The \textbf{Simple Region Crossover (SRC)} concerns grouping all variables that refer to a single urban region represented by one of the crossed individuals (Fig. \ref{fig:depAndCross} left column, middle row). We present an example, in which the grouped variables do not represent continuous region of areas but they are joined in one region by the fixed urban areas. While creating offspring SRC will choose half of the clustered regions from the first parent. The disadvantage of SRC is that it creates relatively large groups of differing genes. Additionally, if two crossed individuals express similar land-use maps, then, using SRC, they will be unable to exchange differing parts of those regions that are similar. Thus, SRC should assure fast convergence by exchanging large regions that refer to high-quality solutions but may be less effective in improving the already-found high quality solutions.\par

To overcome potential SRC disadvantages, in \textbf{Differing Region Crossover (DRC)}, we propose to cluster urban-differing areas (Fig. \ref{fig:depAndCross} right column, middle row). For the exemplary individuals, such crossover would yield three different variable clusters. Following the idea of generating masks that precisely refer to region differences, we also propose the \textbf{Individual-based Differing Region Crossover (IDRC)}. IDRC works in the same way as DRC, but it groups only the variables from the same category. Therefore, in the proposed example (Fig. \ref{fig:depAndCross} right column, bottom row) the second DRC mask is divided into two masks.\par

Individuals resulting from all of the proposed crossover types may be infeasible. After crossover, with a given probability, we execute MutC or MutC2 (depending on the optimizer's configuration). If an individual is infeasible after these operations, then it's repaired with RRM or BRM procedure.

\section{Proposed Initialization Procedures}
\label{sec:inits}

Proposing problem-dedicated mutation operators was successfully aimed in \cite{luoCoreSchwaab}. In the previous section, we show that we can exploit problem-dependent intuitions to propose variable dependency concepts and enable the clusterization of dependent variables even if standard dependency concepts are not applicable. However, the potential of proposing effective problem-dependent mechanisms is not limited to mutation and crossover. Some studies indicate that generating an adequate initial population can improve results quality \cite{luoInit1,luoInit2}. Therefore, we propose four population initialization procedures to verify their potential for improving the results. Some of these procedures are based on the same or similar intuitions employed by the mutation operators presented in Section \ref{sec:problem:StandOperators}.\par

In \textbf{Soil Quality Biased Initialization (SQ-I)}, we create individuals by randomly choosing agricultural areas and transforming them into urban ones with probability $1-sq$, where $sq$ is the normalized soil quality of a given area, which is cell's $a_{rck}$ divided by maximum $a_{rck}$ value for given sample. This operation continues until $T$ areas are transformed. As in SP-I (see Section \ref{sec:problem:StandOperators}), the agricultural areas to transform are chosen randomly, but the areas of high value are chosen with a lower probability. SQI-resulting population shall contain individuals that assure a lower loss of agricultural quality.\par

In \textbf{Total Edge Length Biased Initialization (TEL-I)}, we create individuals by randomly selecting agricultural areas to transform into urban but a given area is transformed with a probiblity of $u/4$, where $u$ is the number of adjacent urban areas. TEL-I should propose populations with individuals that are well-adjusted to TEL measure.\par

The research presented in \cite{luoCoreSchwaab} shows that using various combinations of mutation operators can yield better results by considering both objective functions at the same time. Therefore, we propose the following hybrids of SQ-I and TEL-I.\par

The \textbf{Hybrid SQ/TEL Initialization (HYB-I)}, works in the same way as SQ-I and TEL-I, but the conversion probability is equal to $\sqrt{(1-sq)*(u/4)}$. Thus, HYB-I combines the features of of SQ-I and TEL-I. In \textbf{Half by Half Initialization (HAL-I)}, one half of the population is created using SQ-I and the other half using TEL-I.\par

\section{Experiments}
\label{sec:exp}
In this work, we propose new crossover operators and population initialization procedures dedicated to the considered LUO problem. The objective of the proposed experiments is to answer the following questions:\\
\textbf{Q1.} Can the proposed crossover operators (SRC, DRC, and IDRC) improve the optimizer's effectiveness compared to the standard AC crossover?\\
\textbf{Q2.} Do the proposed population initialization procedures positively impact the quality of the results?\\
\textbf{Q3.} Is the positive or negative influence of the proposed mechanisms optimizer-dependent?\par

To answer the above questions, we introduce the proposed mechanisms into NSGA-II and MOEA/D, well-known MO-dedicated optimizers. Another advantage of this choice is that the procedures of NSGA-II and MOEA/D differ significantly, e.g., the first considers the domination relation to decide which individuals are of higher quality, while the latter scalarizes fitness for this purpose. Note that the prior research presented in \cite{luoCoreSchwaab} considers only NSGA-II.\par

\subsection{Experiment setup}
\label{sec:exp:setup}


We use 14 LUO benchmarks, twelve of which were already considered in \cite{schwaab2017reducing, luoCoreSchwaab}. Three of them are used for tuning and the initial operator influence analysis. The stop condition was $100 000$ fitness function evaluations (FFE) and is the same as in \cite{luoCoreSchwaab}. All optimizers and their versions were implemented in Python. For NSGA-II we used selection mechanism provided by DEAP framework \footnote{\url{github.com/DEAP/deap}}. For MOEA/D, we provided implementation  based on the source code provided at \footnote{\url{sites.google.com/view/moead/home}}.\par

Each optimizer (NSGA-II/MOEA/D) was considered in all available configurations dependent on crossover operator (AC/SRC/DRC/ IDRC), mutation operator (MutC/MutC2), population initialization procedure (SP-I/SQ-i/TEL-I/HYB-I/HAL-I), and the repair operator (RRM, BRM). Based on their results, the most promising operators were selected, for which all possible combinations were then the subject of the tunning procedure that is presented below.\par

Each combination was tunned two times, once with starting parameters of population size 100, crossover and mutation probability 50\%, while the second-time crossover probability 90\% and mutation probability 10\%. To compare parameter setups, 5 random seeds were used due to the stochastic nature of the process.\par

First, the average hypervolume achieved by a combination of initial parameters was computed. Then, the population was tuned. In 10 iterations, starting with a given step (of size 40), we modify the best-found population size (increasing or decreasing it). For the population size analysed in a given iteration, the average hypervolume for a given setup was computed. If the score was better than the original, it was marked as a current best parameter. If it was worse, the direction was changed (addition to subtraction and the opposite). If the change was from subtraction to addition, the step size was divided in half first. After 10 iterations of the process, the current best population became the base parameter for further tuning, as well as the population size and results for the tuning. \par 

The crossover probability, and mutation probability were tuned in the same manner. With an initial adjustment step of 5\%.\par

All steps included guardrails to handle edge cases of invalid parameters of zero or negative population size or probability, as well as probability higher than 100\%. Those however do not change overall idea of the process.\par

For the best combinations, each experiment was repeated 20 times, now using all 14 samples. To verify the statistical significance of the result differences we used the Wilcoxon signed-rank test with 5\% significance level. The complete source code and the complete detailed results of the experiments are available at Zenodo~\cite{zenodo-package} and GitHub\footnote{\url{github.com/KubaMaciazek/Gray-Box-concepts-for-Land-Use-Assignment}}\par

The considered LUO problem is MO. Therefore, we consider two measures to check the quality of the proposed PFs. Inverted Generational Distance (IGD) is defined as follows.
	\begin{equation}
	\label{eq:igd}
	D_{\mathcal{P}_F\to \mathcal{S}}(\mathcal{S}) = \frac{1}{|\mathcal{P}_F|} \sum_{\substack{f^0 \in \mathcal{P}_F}} \min_{x \in \mathcal{S}}{\{d(f(x), f^0) \}},
	\end{equation}
	where $\mathcal{P}_F$ is the optimal PF, $\mathcal{S}$ is the evaluated PF, and $d(\cdot,\cdot)$ is the Euclidean distance. 

The optimal IGD value is $0$, occurring when $\mathcal{S}$ covers $\mathcal{P}_F$. Here, we consider the real-world LUO instances for which $\mathcal{P}_F$ is unknown. Therefore, we construct the pseudo-optimal PF in the following manner. We consider all PFs returned by all optimizers in all runs and choose non-dominated points from such a set. Such a pseudo-optimal PF is not necessarily optimal but is better or equal to all evaluated PFs and is enough for computing IGD. The same IGD pseudo-optimal PF construction procedure was employed in \cite{moP3}.\par

The second PF quality measure was Hypervolume (HV) defined as 
\begin{equation}
	\label{eq:hv}
	HV(\mathcal{S}) = \Lambda(\bigcup_{s \in \mathcal{S}}{\{s' | s \prec s' \prec s^{nadir}\}})
\end{equation}
where $\Lambda$ is a Lebesgue measure, $\mathcal{S}$ is the evaluated PF, $s$ is a point from $\mathcal{S}$, and $s^{nadir}$ is the Nadir Point \cite{MOmeasures}.\par

In the results referring to HV and IGD, we consider their normalized values computed as follows: $N(V_r) = (V_r - V_{min})/(V_{max}-V_{min})$, where $N(V_r)$ is the normalization function, $V_r$ is the normalized value, $V_{min}$, $V_{max}$ are the minimal and maximal values of the given measure obtained by all considered optimizers, respectively. Note that HV is maximized and IGD is minimized. The other considered normalization type is a ranking.

\subsection{Results}

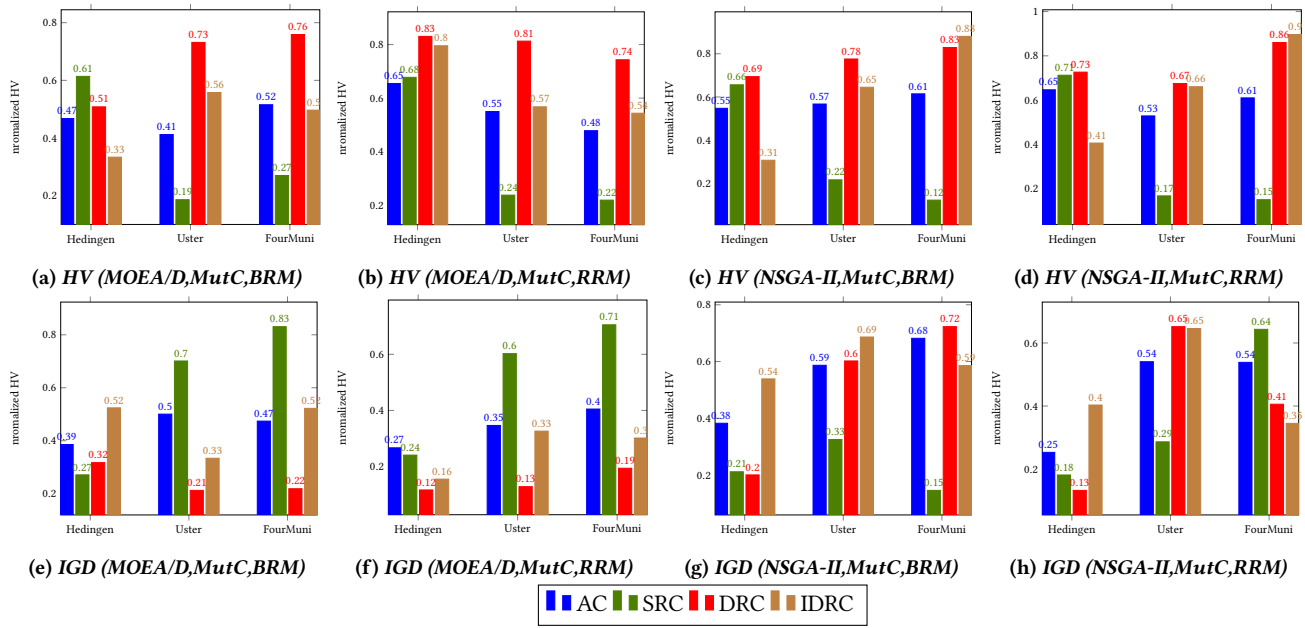
\begin{figure*}[h]
\begin{subfigure}[b]{0.24\linewidth}
\resizebox{\linewidth}{!}{%
\begin{tikzpicture}
\begin{axis}[
    ybar,
    enlargelimits=0.15,
    legend style={at={(0.5,-0.15)},
      anchor=north,legend columns=-1},
    ylabel={nromalized HV},
    symbolic x coords={Hedingen,Uster,FourMuni},
    xtick=data,
    nodes near coords,
    nodes near coords align={vertical},
    legend entries={AC,SRC,DRC,IDRC},
    legend columns=-1,
    legend to name=named,
]
\addplot[blue,fill=blue]
	coordinates {(Hedingen,0.467548993190021) (Uster,0.412073142759738) (FourMuni,0.51543122367615)};

\addplot[darkgreen,fill=darkgreen]
   coordinates {(Hedingen,0.613794429364375) (Uster,0.186050184412624) (FourMuni,0.269941509489442)};

\addplot[red,fill=red]
	coordinates {(Hedingen,0.508103499669465) (Uster,0.731671267591574) (FourMuni,0.758912114233774)};

\addplot[brown,fill=brown]
    coordinates {(Hedingen,0.333658275338349) (Uster,0.557837409289336) (FourMuni,0.496677992059662)};
    
\end{axis}
\end{tikzpicture}
}
\caption{\textit{HV (MOEA/D,MutC,BRM)}}
		\label{fig:operatorInfluence:moeadBRM:HV}
\end{subfigure}
\begin{subfigure}[b]{0.24\linewidth}
\resizebox{\linewidth}{!}{%
\begin{tikzpicture}
\begin{axis}[
    ybar,
    enlargelimits=0.15,
    legend style={at={(0.5,-0.15)},
      anchor=north,legend columns=-1},
    ylabel={nromalized HV},
    symbolic x coords={Hedingen,Uster,FourMuni},
    xtick=data,
    nodes near coords,
    nodes near coords align={vertical},
    legend entries={AC,SRC,DRC,IDRC},
    legend columns=-1,
    legend to name=named,
]
\addplot[blue,fill=blue]
   coordinates {(Hedingen,0.654559877177811) (Uster,0.549815772474007) (FourMuni,0.478215639111512)};

\addplot[darkgreen,fill=darkgreen]
    coordinates {(Hedingen,0.677534536709667) (Uster,0.237410592933938) (FourMuni,0.218825264526664)};

\addplot[red,fill=red]
    coordinates {(Hedingen,0.830518429860242) (Uster,0.812850620274986) (FourMuni,0.743176260632451)};

\addplot[brown,fill=brown]
    coordinates {(Hedingen,0.795420812783436) (Uster,0.567343136126041) (FourMuni,0.543577133157338)};

\end{axis}
\end{tikzpicture}
}
\caption{\textit{HV (MOEA/D,MutC,RRM)}}
		\label{fig:operatorInfluence:moeadRRM:HV}
\end{subfigure}
\begin{subfigure}[b]{0.24\linewidth}
\resizebox{\linewidth}{!}{%
\begin{tikzpicture}
\begin{axis}[
    ybar,
    enlargelimits=0.15,
    legend style={at={(0.5,-0.15)},
      anchor=north,legend columns=-1},
    ylabel={nromalized HV},
    symbolic x coords={Hedingen,Uster,FourMuni},
    xtick=data,
    nodes near coords,
    nodes near coords align={vertical},
    legend entries={AC,SRC,DRC,IDRC},
    legend columns=-1,
    legend to name=named,
]
\addplot[blue,fill=blue]
    coordinates {(Hedingen,0.547619845544215) (Uster,0.567398806528652) (FourMuni,0.614513252336899)};

\addplot[darkgreen,fill=darkgreen]
    coordinates {(Hedingen,0.656311031010514) (Uster,0.216642512236718) (FourMuni,0.122768678741304)};

\addplot[red,fill=red]
   coordinates {(Hedingen,0.694140627270487) (Uster,0.775841707557213) (FourMuni,0.828936254083399)};

\addplot[brown,fill=brown]
    coordinates {(Hedingen,0.307278137953821) (Uster,0.645068699413647) (FourMuni,0.880425790392625)};
    
\end{axis}
\end{tikzpicture}
}
\caption{\textit{HV (NSGA-II,MutC,BRM)}}
		\label{fig:operatorInfluence:nsgaBRM:HV}
\end{subfigure}
\begin{subfigure}[b]{0.24\linewidth}
\resizebox{\linewidth}{!}{%
\begin{tikzpicture}
\begin{axis}[
    ybar,
    enlargelimits=0.15,
    legend style={at={(0.5,-0.15)},
      anchor=north,legend columns=-1},
    ylabel={nromalized HV},
    symbolic x coords={Hedingen,Uster,FourMuni},
    xtick=data,
    nodes near coords,
    nodes near coords align={vertical},
    legend entries={AC,SRC,DRC,IDRC},
    legend columns=-1,
    legend to name=named,
]
\addplot[blue,fill=blue]
   coordinates {(Hedingen,0.646544705660602) (Uster,0.52855952051284) (FourMuni,0.610047370651664)};

\addplot[darkgreen,fill=darkgreen]
    coordinates {(Hedingen,0.712466234275758) (Uster,0.167333847616022) (FourMuni,0.150152796330187)};

\addplot[red,fill=red]
    coordinates {(Hedingen,0.726378957754919) (Uster,0.674749219922171) (FourMuni,0.860183285360501)};

\addplot[brown,fill=brown]
    coordinates {(Hedingen,0.405995552731563) (Uster,0.660480881705867) (FourMuni,0.896196835959038)};

\end{axis}
\end{tikzpicture}
}
\caption{\textit{HV (NSGA-II,MutC,RRM)}}
		\label{fig:operatorInfluence:nsgaRRM:HV}
\end{subfigure}


\begin{subfigure}[b]{0.24\linewidth}
\resizebox{\linewidth}{!}{%
\begin{tikzpicture}
\begin{axis}[
    ybar,
    enlargelimits=0.15,
    legend style={at={(0.5,-0.15)},
      anchor=north,legend columns=-1},
    ylabel={nromalized HV},
    symbolic x coords={Hedingen,Uster,FourMuni},
    xtick=data,
    nodes near coords,
    nodes near coords align={vertical},
    legend entries={AC,SRC,DRC,IDRC},
    legend columns=-1,
    legend to name=named,
]

\addplot[blue,fill=blue]
	coordinates {(Hedingen,0.38566372197342) (Uster,0.500455336811833) (FourMuni,0.474084039818467)};

\addplot[darkgreen,fill=darkgreen]
coordinates {(Hedingen,0.271090388956687) (Uster,0.701122141259482) (FourMuni,0.830827546755127)};
    
\addplot[red,fill=red]
	coordinates {(Hedingen,0.31725444893758) (Uster,0.212461460743535) (FourMuni,0.218717520090514)};

\addplot[brown,fill=brown]
    coordinates {(Hedingen,0.52415434061232) (Uster,0.333794143029979) (FourMuni,0.522226840646107)};

\end{axis}
\end{tikzpicture}
}
\caption{\textit{IGD (MOEA/D,MutC,BRM)}}
		\label{fig:operatorInfluence:moeadBRM:IGD}
\end{subfigure}
\begin{subfigure}[b]{0.24\linewidth}
\resizebox{\linewidth}{!}{%
\begin{tikzpicture}
\begin{axis}[
    ybar,
    enlargelimits=0.15,
    legend style={at={(0.5,-0.15)},
      anchor=north,legend columns=-1},
    ylabel={nromalized HV},
    symbolic x coords={Hedingen,Uster,FourMuni},
    xtick=data,
    nodes near coords,
    nodes near coords align={vertical},
    legend entries={AC,SRC,DRC,IDRC},
    legend columns=-1,
    legend to name=named,
]

\addplot[blue,fill=blue]
    coordinates {(Hedingen,0.266866843841107) (Uster,0.346560728473786) (FourMuni,0.404995220463773)};

\addplot[darkgreen,fill=darkgreen]
    coordinates {(Hedingen,0.240917022634269) (Uster,0.602311131227762) (FourMuni,0.705082403108459)};
    
\addplot[red,fill=red]
    coordinates {(Hedingen,0.116737970506164) (Uster,0.128542174552511) (FourMuni,0.194134107803426)};

\addplot[brown,fill=brown]
    coordinates {(Hedingen,0.155411648197786) (Uster,0.326375538920984) (FourMuni,0.301761999590137)};

\end{axis}
\end{tikzpicture}
}
\caption{\textit{IGD (MOEA/D,MutC,RRM)}}
		\label{fig:operatorInfluence:moeadRRM:IGD}
\end{subfigure}
\begin{subfigure}[b]{0.24\linewidth}
\resizebox{\linewidth}{!}{%
\begin{tikzpicture}
\begin{axis}[
    ybar,
    enlargelimits=0.15,
    legend style={at={(0.5,-0.15)},
      anchor=north,legend columns=-1},
    ylabel={nromalized HV},
    symbolic x coords={Hedingen,Uster,FourMuni},
    xtick=data,
    nodes near coords,
    nodes near coords align={vertical},
    legend entries={AC,SRC,DRC,IDRC},
    legend columns=-1,
    legend to name=named,
]

\addplot[blue,fill=blue]
    coordinates {(Hedingen,0.382541670048373) (Uster,0.586839815210613) (FourMuni,0.681900582393319)};

\addplot[darkgreen,fill=darkgreen]
    coordinates {(Hedingen,0.211545235076937) (Uster,0.325124922855824) (FourMuni,0.146010348827088)};   
    
\addplot[red,fill=red]
    coordinates {(Hedingen,0.200535557076638) (Uster,0.602095100791387) (FourMuni,0.722985632212698)};

\addplot[brown,fill=brown]
    coordinates {(Hedingen,0.538880324599495) (Uster,0.686463725816799) (FourMuni,0.585657800669198)};

\end{axis}
\end{tikzpicture}
}
\caption{\textit{IGD (NSGA-II,MutC,BRM)}}
		\label{fig:operatorInfluence:nsga2BRM:IGD}
\end{subfigure}
\begin{subfigure}[b]{0.24\linewidth}
\resizebox{\linewidth}{!}{%
\begin{tikzpicture}
\begin{axis}[
    ybar,
    enlargelimits=0.15,
    legend style={at={(0.5,-0.15)},
      anchor=north,legend columns=-1},
    ylabel={nromalized HV},
    symbolic x coords={Hedingen,Uster,FourMuni},
    xtick=data,
    nodes near coords,
    nodes near coords align={vertical},
    legend entries={AC,SRC,DRC,IDRC},
    legend columns=-1,
    legend to name=named,
]

\addplot[blue,fill=blue]
    coordinates {(Hedingen,0.253012451529119) (Uster,0.540617852162089) (FourMuni,0.538515462305613)};

\addplot[darkgreen,fill=darkgreen]
    coordinates {(Hedingen,0.181235698740671) (Uster,0.28639294324486) (FourMuni,0.643351727382637)};
    
\addplot[red,fill=red]
    coordinates {(Hedingen,0.132361743032068) (Uster,0.651864514070768) (FourMuni,0.40572966527033)};

\addplot[brown,fill=brown]
    coordinates {(Hedingen,0.403230781178677) (Uster,0.645851342614866) (FourMuni,0.345003792409422)};

\end{axis}
\end{tikzpicture}
}
\caption{\textit{IGD (NSGA-II,MutC,RRM)}}
		\label{fig:operatorInfluence:nsga2RRM:IGD}
\end{subfigure}

        \hspace{1.11 cm}
	\ref{named}
	\caption{The influence of the proposed crossover operators (base optimizer, mutation type, crossover results repair operation)}
	\label{fig:operatorInfluence}
\end{figure*}

\begin{table*}[]
\caption{The general ranking-based comparison of the best optimizers and their baseline versions}
\label{tab:res:generalRank}
\scriptsize
\begin{tabular}{lllll|lll|c|cc|cccc|cccc}
\multicolumn{5}{c}{\textbf{Optimizer version}} & \multicolumn{3}{c}{\textbf{Configuration}}     &  \textbf{Joined rank}  & \multicolumn{2}{c}{\textbf{vs. Base}}       & \multicolumn{4}{c}{\textbf{IGD ranking}} & \multicolumn{4}{c}{\textbf{HV ranking}}  \\
\textbf{Opt}     & \textbf{Init}  & \textbf{Mut}   & \textbf{Cross} & \textbf{Repair} & \textbf{Pop.} & \textbf{Cross} & \textbf{Mut} & \textbf{(IGD+HV)/2} & \textbf{Better} & \textbf{Worse} & \textbf{Med} & \textbf{Avr}  & \textbf{Min} & \textbf{Max} & \textbf{Med} & \textbf{Avr}  & \textbf{Min} & \textbf{Max} \\
\hline
MOEA/D  & HYB-I & MutC  & DRC & RRM   & 100     & 0.5    & 0.5  & 2.29         & 14     & 0     & 1.5 & 2.21 & 1   & 6   & 2   & 2.36 & 1   & 6   \\
MOEA/D  & TEL-I & MutC2 & DRC & RRM  & 100     & 0.5    & 0.5  & 2.46         & 13     & 1    & 2   & 2.71 & 1   & 8   & 2   & 2.21 & 1   & 7   \\
MOEA/D  & SQ-I  & MutC2 & DRC & RRM  & 110     & 0.5    & 0.6  & 2.82         & 14     & 0     & 2.5 & 2.71 & 1   & 4   & 3   & 2.93 & 1   & 4   \\
MOEA/D  & SQ-I  & MutC2 & DRC & RRM  & 100     & 0.5    & 0.5  & 3.68         & 13     & 1    & 3   & 3.64 & 3   & 6   & 4   & 3.71 & 1   & 6   \\
NSGA-II & HAL-I & MutC  & IDRC & BRM & 340     & 1.0    & 0.1  & 5.18         & 14     & 0     & 6   & 5.43 & 1   & 8   & 5.5 & 4.93 & 1   & 6   \\
MOEA/D  & SP-I  & MutC  & AC  & RRM  & 100     & 0.5    & 1.0  & 5.75         & \multicolumn{2}{c|}{Base MOEA/D}  & 5   & 5.71 & 4   & 8   & 5   & 5.79 & 3   & 9   \\
NSGA-II & HAL-I & MutC2 & DRC & BRM  & 300     & 1.0    & 0.5  & 6.43         & 14     & 0     & 6.5 & 6.29 & 2   & 9   & 7   & 6.57 & 3   & 8   \\
NSGA-II & SQ-I  & MutC2 & DRC & RRM  & 120     & 0.9    & 0.5  & 8.11         & 11      & 3    & 8   & 8.07 & 6   & 10  & 8   & 8.14 & 7   & 9   \\
NSGA-II & HAL-I & MutC2 & DRC & RRM  & 100     & 0.5    & 0.5  & 8.75         & 6     & 8    & 10  & 9.00 & 5   & 10  & 9   & 8.50 & 5   & 10  \\
NSGA-II & SP-I  & MutC  & AC  & RRM  & 100     & 0.5    & 1.0  & 9.54         & \multicolumn{2}{c|}{Base NSGA-II}     & 9   & 9.21 & 8   & 10  & 10  & 9.86 & 9   & 10 
\end{tabular}
\end{table*}

In Sections \ref{sec:operators} and \ref{sec:inits}, we have proposed a series of problem-dedicated mechanisms. However, the key proposition of this work is three proposed crossovers (SRC, DRC, IDRC) based on the proposed concepts of variable dependency understanding LUO. Their objective was to support effective mixing since gray-box mechanisms are unavailable. Therefore, we wish to compare the base MOEA/D and NSGA-II versions and their versions using one of the three proposed crossover operators. Base versions use the standard operators described in Section \ref{sec:problem:StandOperators}, i.e., SP-I for population initialization, AC for crossover, and MutC for mutation. \par

In Fig. \ref{fig:operatorInfluence}, we show the detailed comparison of the three LUO instances considered in the tuning procedure. We report the HV and IGD-based comparison between base MOEA/D and NSGA-II optimizers and their versions using one of the three proposed crossover operators (with RRM or BRM employed to repair the crossover result). Fig. \ref{fig:operatorInfluence} shows that using DRC outperforms AC for every considered test case for both optimizers. IDRC also outperforms AC but not in all cases. SRC performs worse than AC. For IGD and MOEA/D DRC outperforms AC in all cases, too. However, for IGD and NSGA-II DRC and AC perform equally well. The \textit{p}-values for the considered comparisons can be found in the supplementary material.\par

The above observations allow for the following initial conclusions. SRC creates too large masks because it considers too many irrelevant dependencies. Considering the typical linkage-based concepts, we can state that SRC suffers from \textit{false linkage} \cite{3lo}, i.e., considers some of the variables dependent, although they are independent. Therefore, it creates too large masks, which may deteriorate the quality of the result \cite{linkageQuality}). Therefore, its performance is low when compared to other crossover types.\par

DRC is the most effective for all considered MOEA/D and NSGA-II versions. IDRC performs worse than DRC. Such observation shows that IDRC creates too small masks and ignores too many dependencies between variables. Thus, we can say that it suffers from \textit{missing linkage}, i.e., ignores some of the dependencies between variables, which results in masks that are too small and deteriorate the quality of the results \cite{missingIsBad}.\par

In the main part of the experiments, we have considered all possible optimizer versions and all 14 LUO instances. For practical reasons, we report the results for the four best versions and the base version for each of the two considered optimizers. Complete results for all optimizer versions are available in the source code, and results pack pointed out in Section \ref{sec:exp:setup}. Table \ref{tab:res:generalRank} presents the detailed configurations of the five chosen versions for each optimizer and their IGD- and HV-based rankings. The optimizers are sorted considering their quality indicated by the \textit{joined ranking} (average of IGD- and HV-based rankings). Additionally, we report the number of considered LUO instances for which a given version was better or worse than its base version.\par

Table \ref{tab:res:generalRank} is coherent with the results presented in Fig. \ref{fig:operatorInfluence}. All most effective MOEA/D versions and three on four NSGA-II versions use DRC crossover. The three best MOEA/D versions, with similar ranking results, consider three different initialization procedures (including the simplest SQ-I), showing that their influence on the result quality is negligible. Two of them use MutC2 for mutation, and one MutC, which confirms their influence on the quality of the results, is low, too. No matter which MOEA/D version is considered, their tuned parameter settings are either the same or similar. Thus, the parameter settings did not influence the results significantly. Finally, all four best MOEA/D versions outperformed the base MOEA/D version for all 14 considered instances (two versions) or for 13 instances, which confirms that the proposed DRC crossover brings the effectiveness improvement potential similar to typical gray-box operators for problems to which they are applicable \cite{whitleyNext}.\par

DRC crossover increased the effectiveness of NSGA-II, too. The two best NSGA-II versions that use IDRC (the only optimizer version on the list that uses a different crossover than DRC) outperform the base NSGA-II version for all considered cases. This result remains convincing for the third-best NSGA-II version that performs better for 11 instances. However, the third-best version was outperformed by the base NSGA-II. Such results show that the proposed crossover positively influences the effectiveness of NSGA-II. Note that, in general, NSGA-II performs significantly worse than MOEA/D. Such a result is not a surprise since similar observations can be made based on other state-of-the-art research \cite{moeadStrong,moeadStrong2,moeadStrongIGD}. In addition to the above results, we support the Pareto front visualization in the supplementary material.\par

The results confirm that the proposed DRC crossover significantly improves the effectiveness of the two considered MO optimizers. Together with other mechanisms proposed in this paper, DRC allows for proposing a LUO-dedicated optimizer that significantly outperforms the considered competing baseline optimizers.

\section{Conclusions}
\label{sec:conc}

In this work, we consider a real-world MO NP-hard problem. Since the considered LUO setting is gray-box elusive, we propose problem-dedicated concepts of variable dependency. Using them we propose the DRC crossover. The experiments show that DRC significantly improves the effectiveness of the two well-know MO-dedicated optimizers based on significantly different concepts. Together with other problem dedicated mechanisms, we propose new MOEA/D- and NSGA-II-based optimizers that outperform the competing optimizers for all considered LUO instances.\par

The main future work directions shall focus on proposing the problem-dedicated parameter-less optimizers for LUO \cite{moP3,MoGomeaSwarm}. The other natural option is proposing the donor-based mixing operator, similar to PX \cite{pxForBinary} or Optimal Mixing \cite{ltga}. Finally, an interesting idea is to propose a LUO representation that would not be gray-box-elusive and allow using the gray-box leverage. The main difficulty in such a research track is obtaining a problem representation for which all or almost all variables are interdependent, which would make gray-box operators non-applicable \cite{wVIG}. Finally, considering the dependencies predicted by the statistical analysis \cite{ltga,dsmga2} seems highly promissing.

\begin{acks}
  Jakub Maciazek and Michal Przewozniczek were supported by the Polish National Science Centre (NCN) under grant number 2020/38/E/ST6/00370. Jonas Schwaab was supported by the Swiss National Science Foundation (SNSF) through the Scientific Exchanges grant IZSEZ0\_229824.
\end{acks}

%% file: landShare.bbl

\begin{thebibliography}{48}


\ifx \showCODEN    \undefined \def \showCODEN     #1{\unskip}     \fi
\ifx \showISBNx    \undefined \def \showISBNx     #1{\unskip}     \fi
\ifx \showISBNxiii \undefined \def \showISBNxiii  #1{\unskip}     \fi
\ifx \showISSN     \undefined \def \showISSN      #1{\unskip}     \fi
\ifx \showLCCN     \undefined \def \showLCCN      #1{\unskip}     \fi
\ifx \shownote     \undefined \def \shownote      #1{#1}          \fi
\ifx \showarticletitle \undefined \def \showarticletitle #1{#1}   \fi
\ifx \showURL      \undefined \def \showURL       {\relax}        \fi
\providecommand\bibfield[2]{#2}
\providecommand\bibinfo[2]{#2}
\providecommand\natexlab[1]{#1}
\providecommand\showeprint[2][]{arXiv:#2}

\bibitem[Aerts et~al\mbox{.}(2003)]%
        {luoNotation}
\bibfield{author}{\bibinfo{person}{Jeroen C. J.~H. Aerts}, \bibinfo{person}{Erwin Eisinger}, \bibinfo{person}{Gerard B.~M. Heuvelink}, {and} \bibinfo{person}{Theodor~J. Stewart}.} \bibinfo{year}{2003}\natexlab{}.
\newblock \showarticletitle{Using Linear Integer Programming for Multi-Site Land-Use Allocation}.
\newblock \bibinfo{journal}{\emph{Geographical Analysis}} \bibinfo{volume}{35}, \bibinfo{number}{2} (\bibinfo{year}{2003}), \bibinfo{pages}{148--169}.
\newblock
\href{https://doi.org/10.1111/j.1538-4632.2003.tb01106.x}{doi:\nolinkurl{10.1111/j.1538-4632.2003.tb01106.x}}
\showeprint{https://onlinelibrary.wiley.com/doi/pdf/10.1111/j.1538-4632.2003.tb01106.x}


\bibitem[Bennett et~al\mbox{.}(2004)]%
        {luoInit1}
\bibfield{author}{\bibinfo{person}{David~A. Bennett}, \bibinfo{person}{Ningchuan Xiao}, {and} \bibinfo{person}{Marc~P. Armstrong}.} \bibinfo{year}{2004}\natexlab{}.
\newblock \showarticletitle{Exploring the Geographic Consequences of Public Policies Using Evolutionary Algorithms}.
\newblock \bibinfo{journal}{\emph{Annals of the Association of American Geographers}} \bibinfo{volume}{94}, \bibinfo{number}{4} (\bibinfo{year}{2004}), \bibinfo{pages}{827--847}.
\newblock
\href{https://doi.org/10.1111/j.1467-8306.2004.00437.x}{doi:\nolinkurl{10.1111/j.1467-8306.2004.00437.x}}
\showeprint{https://onlinelibrary.wiley.com/doi/pdf/10.1111/j.1467-8306.2004.00437.x}


\bibitem[Chikumbo et~al\mbox{.}(2014)]%
        {envirChikumbo}
\bibfield{author}{\bibinfo{person}{Oliver Chikumbo}, \bibinfo{person}{Erik Goodman}, {and} \bibinfo{person}{Kalyanmoy Deb}.} \bibinfo{year}{2014}\natexlab{}.
\newblock \showarticletitle{Triple Bottomline Many-Objective-Based Decision Making for a Land Use Management Problem}.
\newblock \bibinfo{journal}{\emph{Journal of Multi-Criteria Decision Analysis}}  \bibinfo{volume}{22} (\bibinfo{date}{12} \bibinfo{year}{2014}).
\newblock
\href{https://doi.org/10.1002/mcda.1536}{doi:\nolinkurl{10.1002/mcda.1536}}


\bibitem[Deb and Goldberg(1993)]%
        {decFunc}
\bibfield{author}{\bibinfo{person}{Kalyanmoy Deb} {and} \bibinfo{person}{David~E. Goldberg}.} \bibinfo{year}{1993}\natexlab{}.
\newblock \showarticletitle{Sufficient Conditions for Deceptive and Easy Binary Functions}.
\newblock \bibinfo{journal}{\emph{Ann. Math. Artif. Intell.}} \bibinfo{volume}{10}, \bibinfo{number}{4} (\bibinfo{year}{1993}), \bibinfo{pages}{385--408}.
\newblock


\bibitem[Deb et~al\mbox{.}(2002)]%
        {nsga2}
\bibfield{author}{\bibinfo{person}{K. Deb}, \bibinfo{person}{A. Pratap}, \bibinfo{person}{S. Agarwal}, {and} \bibinfo{person}{T. Meyarivan}.} \bibinfo{year}{2002}\natexlab{}.
\newblock \showarticletitle{A fast and elitist multiobjective genetic algorithm: NSGA-II}.
\newblock \bibinfo{journal}{\emph{IEEE Transactions on Evolutionary Computation}} \bibinfo{volume}{6}, \bibinfo{number}{2} (\bibinfo{date}{April} \bibinfo{year}{2002}), \bibinfo{pages}{182--197}.
\newblock
\showISSN{1089-778X}


\bibitem[Derbel et~al\mbox{.}(2021)]%
        {moeadStrong2}
\bibfield{author}{\bibinfo{person}{Bilel Derbel}, \bibinfo{person}{Geoffrey Pruvost}, {and} \bibinfo{person}{Byung-Woo Hong}.} \bibinfo{year}{2021}\natexlab{}.
\newblock \showarticletitle{Enhancing Moea/d with Escape Mechanisms}. In \bibinfo{booktitle}{\emph{2021 IEEE Congress on Evolutionary Computation (CEC)}} (Krak\'{o}w, Poland). \bibinfo{publisher}{IEEE Press}, \bibinfo{pages}{1163–1170}.
\newblock
\href{https://doi.org/10.1109/CEC45853.2021.9504957}{doi:\nolinkurl{10.1109/CEC45853.2021.9504957}}


\bibitem[Dushatskiy et~al\mbox{.}(2021)]%
        {ellGomea}
\bibfield{author}{\bibinfo{person}{Arkadiy Dushatskiy}, \bibinfo{person}{Tanja Alderliesten}, {and} \bibinfo{person}{Peter A.~N. Bosman}.} \bibinfo{year}{2021}\natexlab{}.
\newblock \showarticletitle{A Novel Approach to Designing Surrogate-assisted Genetic Algorithms by Combining Efficient Learning of Walsh Coefficients and Dependencies}.
\newblock \bibinfo{journal}{\emph{ACM Trans. Evol. Learn. Optim.}} \bibinfo{volume}{1}, \bibinfo{number}{2}, Article \bibinfo{articleno}{5} (\bibinfo{date}{July} \bibinfo{year}{2021}), \bibinfo{numpages}{23}~pages.
\newblock
\href{https://doi.org/10.1145/3453141}{doi:\nolinkurl{10.1145/3453141}}


\bibitem[Heckendorn(2002)]%
        {heckendorn2002}
\bibfield{author}{\bibinfo{person}{R.~B. Heckendorn}.} \bibinfo{year}{2002}\natexlab{}.
\newblock \showarticletitle{Embedded Landscapes}.
\newblock \bibinfo{journal}{\emph{Evolutionary Computation}} \bibinfo{volume}{10}, \bibinfo{number}{4} (\bibinfo{year}{2002}), \bibinfo{pages}{345--369}.
\newblock


\bibitem[Hsu and Yu(2015)]%
        {dsmga2}
\bibfield{author}{\bibinfo{person}{Shih-Huan Hsu} {and} \bibinfo{person}{Tian-Li Yu}.} \bibinfo{year}{2015}\natexlab{}.
\newblock \showarticletitle{Optimization by Pairwise Linkage Detection, Incremental Linkage Set, and Restricted / Back Mixing: {DSMGA}-{II}}. In \bibinfo{booktitle}{\emph{Proceedings of the 2015 Annual Conference on Genetic and Evolutionary Computation}} \emph{(\bibinfo{series}{GECCO '15})}. \bibinfo{publisher}{ACM}, \bibinfo{pages}{519--526}.
\newblock


\bibitem[Huang et~al\mbox{.}(2012)]%
        {LUOnpHard}
\bibfield{author}{\bibinfo{person}{Kangning Huang}, \bibinfo{person}{Xiaoping Liu}, \bibinfo{person}{Xia Li}, \bibinfo{person}{Jiayong Liang}, {and} \bibinfo{person}{Shenjing He}.} \bibinfo{year}{2012}\natexlab{}.
\newblock \showarticletitle{An improved artificial immune system for seeking the Pareto front of land use allocation problem in large areas}.
\newblock \bibinfo{journal}{\emph{International Journal of Geographical Information Science}}  \bibinfo{volume}{27} (\bibinfo{date}{11} \bibinfo{year}{2012}), \bibinfo{pages}{922--946}.
\newblock
\href{https://doi.org/10.1080/13658816.2012.730147}{doi:\nolinkurl{10.1080/13658816.2012.730147}}


\bibitem[Kaim et~al\mbox{.}(2018)]%
        {envirKaim}
\bibfield{author}{\bibinfo{person}{Andrea Kaim}, \bibinfo{person}{Anna Cord}, {and} \bibinfo{person}{Martin Volk}.} \bibinfo{year}{2018}\natexlab{}.
\newblock \showarticletitle{A review of multi-criteria optimization techniques for agricultural land use allocation}.
\newblock \bibinfo{journal}{\emph{Environmental Modelling and Software}}  \bibinfo{volume}{105} (\bibinfo{date}{07} \bibinfo{year}{2018}).
\newblock
\href{https://doi.org/10.1016/j.envsoft.2018.03.031}{doi:\nolinkurl{10.1016/j.envsoft.2018.03.031}}


\bibitem[Komarnicki et~al\mbox{.}(2023)]%
        {irrg}
\bibfield{author}{\bibinfo{person}{Marcin~Michal Komarnicki}, \bibinfo{person}{Michal~Witold Przewozniczek}, \bibinfo{person}{Halina Kwasnicka}, {and} \bibinfo{person}{Krzysztof Walkowiak}.} \bibinfo{year}{2023}\natexlab{}.
\newblock \showarticletitle{Incremental Recursive Ranking Grouping for Large-Scale Global Optimization}.
\newblock \bibinfo{journal}{\emph{IEEE Transactions on Evolutionary Computation}} \bibinfo{volume}{27}, \bibinfo{number}{5} (\bibinfo{year}{2023}), \bibinfo{pages}{1498--1513}.
\newblock


\bibitem[Laszczyk and Myszkowski(2019)]%
        {MOmeasures}
\bibfield{author}{\bibinfo{person}{Maciej Laszczyk} {and} \bibinfo{person}{Paweł~B. Myszkowski}.} \bibinfo{year}{2019}\natexlab{}.
\newblock \showarticletitle{Survey of quality measures for multi-objective optimization: Construction of complementary set of multi-objective quality measures}.
\newblock \bibinfo{journal}{\emph{Swarm and Evolutionary Computation}}  \bibinfo{volume}{48} (\bibinfo{year}{2019}), \bibinfo{pages}{109--133}.
\newblock
\showISSN{2210-6502}
\href{https://doi.org/10.1016/j.swevo.2019.04.001}{doi:\nolinkurl{10.1016/j.swevo.2019.04.001}}


\bibitem[Lautenbach et~al\mbox{.}(2013)]%
        {luoInit2}
\bibfield{author}{\bibinfo{person}{Sven Lautenbach}, \bibinfo{person}{Martin Volk}, \bibinfo{person}{Michael Strauch}, \bibinfo{person}{Gerald Whittaker}, {and} \bibinfo{person}{Ralf Seppelt}.} \bibinfo{year}{2013}\natexlab{}.
\newblock \showarticletitle{Optimization-based trade-off analysis of biodiesel crop production for managing an agricultural catchment}.
\newblock \bibinfo{journal}{\emph{Environ. Model. Softw.}}  \bibinfo{volume}{48} (\bibinfo{year}{2013}), \bibinfo{pages}{98--112}.
\newblock
\urldef\tempurl%
\url{https://api.semanticscholar.org/CorpusID:9210561}
\showURL{%
\tempurl}


\bibitem[Lepr\^{e}tre et~al\mbox{.}(2019)]%
        {walshSurBin2}
\bibfield{author}{\bibinfo{person}{Florian Lepr\^{e}tre}, \bibinfo{person}{S\'{e}bastien Verel}, \bibinfo{person}{Cyril Fonlupt}, {and} \bibinfo{person}{Virginie Marion}.} \bibinfo{year}{2019}\natexlab{}.
\newblock \showarticletitle{Walsh functions as surrogate model for pseudo-boolean optimization problems}. In \bibinfo{booktitle}{\emph{Proceedings of the Genetic and Evolutionary Computation Conference}} (Prague, Czech Republic) \emph{(\bibinfo{series}{GECCO '19})}. \bibinfo{publisher}{Association for Computing Machinery}, \bibinfo{address}{New York, NY, USA}, \bibinfo{pages}{303–311}.
\newblock
\showISBNx{9781450361118}
\href{https://doi.org/10.1145/3321707.3321800}{doi:\nolinkurl{10.1145/3321707.3321800}}


\bibitem[Luong et~al\mbox{.}(2018)]%
        {MoGomeaSwarm}
\bibfield{author}{\bibinfo{person}{Ngoc~Hoang Luong}, \bibinfo{person}{Han~La Poutré}, {and} \bibinfo{person}{Peter~A.N. Bosman}.} \bibinfo{year}{2018}\natexlab{}.
\newblock \showarticletitle{Multi-objective Gene-pool Optimal Mixing Evolutionary Algorithm with the Interleaved Multi-start Scheme}.
\newblock \bibinfo{journal}{\emph{Swarm and Evolutionary Computation}}  \bibinfo{volume}{40} (\bibinfo{year}{2018}), \bibinfo{pages}{238 -- 254}.
\newblock
\showISSN{2210-6502}
\href{https://doi.org/10.1016/j.swevo.2018.02.005}{doi:\nolinkurl{10.1016/j.swevo.2018.02.005}}


\bibitem[Maciążek et~al\mbox{.}(2025)]%
        {zenodo-package}
\bibfield{author}{\bibinfo{person}{Jakub Maciążek}, \bibinfo{person}{Michal~Witold Przewozniczek}, {and} \bibinfo{person}{Jonas Schwaab}.} \bibinfo{year}{2025}\natexlab{}.
\newblock \bibinfo{title}{Replication package of the publication ``Seeking and leveraging alternative variable dependency concepts in gray-box-elusive bimodal land-use allocation problems''}.
\newblock \bibinfo{howpublished}{Zenodo}.
\newblock
\href{https://doi.org/doi.org/10.5281/zenodo.15198495}{doi:\nolinkurl{doi.org/10.5281/zenodo.15198495}}


\bibitem[Masoumi et~al\mbox{.}(2016)]%
        {LUOnpHard2}
\bibfield{author}{\bibinfo{person}{Zohreh Masoumi}, \bibinfo{person}{Jamshid Maleki}, \bibinfo{person}{Saadi Mesgari}, {and} \bibinfo{person}{Ali Mansourian}.} \bibinfo{year}{2016}\natexlab{}.
\newblock \showarticletitle{Using an Evolutionary Algorithm in Multiobjective Geographic Analysis for Land Use Allocation and Decision Supporting: Land Use Allocation and Decision Supporting}.
\newblock \bibinfo{journal}{\emph{Geographical Analysis}}  \bibinfo{volume}{49} (\bibinfo{date}{06} \bibinfo{year}{2016}).
\newblock
\href{https://doi.org/10.1111/gean.12111}{doi:\nolinkurl{10.1111/gean.12111}}


\bibitem[Munetomo and Goldberg(1999a)]%
        {linc}
\bibfield{author}{\bibinfo{person}{M. Munetomo} {and} \bibinfo{person}{D.E. Goldberg}.} \bibinfo{year}{1999}\natexlab{a}.
\newblock \showarticletitle{A genetic algorithm using linkage identification by nonlinearity check}. In \bibinfo{booktitle}{\emph{IEEE SMC'99 Conference Proceedings. 1999 IEEE International Conference on Systems, Man, and Cybernetics (Cat. No.99CH37028)}}, Vol.~\bibinfo{volume}{1}. \bibinfo{pages}{595--600 vol.1}.
\newblock
\href{https://doi.org/10.1109/ICSMC.1999.814159}{doi:\nolinkurl{10.1109/ICSMC.1999.814159}}


\bibitem[Munetomo and Goldberg(1999b)]%
        {GoldMonotonicity}
\bibfield{author}{\bibinfo{person}{Masaharu Munetomo} {and} \bibinfo{person}{David~E. Goldberg}.} \bibinfo{year}{1999}\natexlab{b}.
\newblock \showarticletitle{Linkage identification by non-monotonicity detection for overlapping functions}.
\newblock \bibinfo{journal}{\emph{Evol. Comput.}} \bibinfo{volume}{7}, \bibinfo{number}{4} (\bibinfo{date}{dec} \bibinfo{year}{1999}), \bibinfo{pages}{377–398}.
\newblock
\showISSN{1063-6560}
\href{https://doi.org/10.1162/evco.1999.7.4.377}{doi:\nolinkurl{10.1162/evco.1999.7.4.377}}


\bibitem[Pal and Bandyopadhyay(2016)]%
        {pfQuality}
\bibfield{author}{\bibinfo{person}{Monalisa Pal} {and} \bibinfo{person}{Sanghamitra Bandyopadhyay}.} \bibinfo{year}{2016}\natexlab{}.
\newblock \showarticletitle{Reliability of convergence metric and hypervolume indicator for many-objective optimization}. In \bibinfo{booktitle}{\emph{2016 2nd International Conference on Control, Instrumentation, Energy \& Communication (CIEC)}}. \bibinfo{pages}{511--515}.
\newblock
\href{https://doi.org/10.1109/CIEC.2016.7513806}{doi:\nolinkurl{10.1109/CIEC.2016.7513806}}


\bibitem[Przewozniczek et~al\mbox{.}(2021a)]%
        {moP3}
\bibfield{author}{\bibinfo{person}{Michal~Witold Przewozniczek}, \bibinfo{person}{Piotr Dziurzanski}, \bibinfo{person}{Shuai Zhao}, {and} \bibinfo{person}{Leandro~Soares Indrusiak}.} \bibinfo{year}{2021}\natexlab{a}.
\newblock \showarticletitle{Multi-Objective parameter-less population pyramid for solving industrial process planning problems}.
\newblock \bibinfo{journal}{\emph{Swarm and Evolutionary Computation}}  \bibinfo{volume}{60} (\bibinfo{year}{2021}), \bibinfo{pages}{100773}.
\newblock
\showISSN{2210-6502}
\href{https://doi.org/10.1016/j.swevo.2020.100773}{doi:\nolinkurl{10.1016/j.swevo.2020.100773}}


\bibitem[Przewozniczek et~al\mbox{.}(2020)]%
        {linkageQuality}
\bibfield{author}{\bibinfo{person}{Michal~W. Przewozniczek}, \bibinfo{person}{Bartosz Frej}, {and} \bibinfo{person}{Marcin~M. Komarnicki}.} \bibinfo{year}{2020}\natexlab{}.
\newblock \showarticletitle{On Measuring and Improving the Quality of Linkage Learning in Modern Evolutionary Algorithms Applied to Solve Partially Additively Separable Problems}. In \bibinfo{booktitle}{\emph{Proceedings of the 2020 Genetic and Evolutionary Computation Conference}} (Canc\'{u}n, Mexico) \emph{(\bibinfo{series}{GECCO '20})}. \bibinfo{publisher}{Association for Computing Machinery}, \bibinfo{address}{New York, NY, USA}, \bibinfo{pages}{742–750}.
\newblock
\showISBNx{9781450371285}


\bibitem[Przewozniczek et~al\mbox{.}(2024)]%
        {2dled}
\bibfield{author}{\bibinfo{person}{Michal~Witold Przewozniczek}, \bibinfo{person}{Bartosz Frej}, {and} \bibinfo{person}{Marcin~Michal Komarnicki}.} \bibinfo{year}{2024}\natexlab{}.
\newblock \showarticletitle{From Direct to Directional Variable Dependencies – Non-Symmetrical Dependencies Discovery in Real-World and Theoretical Problems}.
\newblock \bibinfo{journal}{\emph{IEEE Transactions on Evolutionary Computation}} (\bibinfo{year}{2024}), \bibinfo{pages}{1--1}.
\newblock
\href{https://doi.org/10.1109/TEVC.2024.3496193}{doi:\nolinkurl{10.1109/TEVC.2024.3496193}}


\bibitem[Przewozniczek and Komarnicki(2020)]%
        {3lo}
\bibfield{author}{\bibinfo{person}{Michal~W. Przewozniczek} {and} \bibinfo{person}{Marcin~M. Komarnicki}.} \bibinfo{year}{2020}\natexlab{}.
\newblock \showarticletitle{Empirical Linkage Learning}.
\newblock \bibinfo{journal}{\emph{IEEE Transactions on Evolutionary Computation}} \bibinfo{volume}{24}, \bibinfo{number}{6} (\bibinfo{date}{Dec} \bibinfo{year}{2020}), \bibinfo{pages}{1097--1111}.
\newblock
\showISSN{1941-0026}


\bibitem[Przewozniczek et~al\mbox{.}(2021b)]%
        {dled}
\bibfield{author}{\bibinfo{person}{Michal~W. Przewozniczek}, \bibinfo{person}{Marcin~M. Komarnicki}, {and} \bibinfo{person}{Bartosz Frej}.} \bibinfo{year}{2021}\natexlab{b}.
\newblock \showarticletitle{Direct Linkage Discovery with Empirical Linkage Learning}. In \bibinfo{booktitle}{\emph{Proceedings of the Genetic and Evolutionary Computation Conference}} (Lille, France) \emph{(\bibinfo{series}{GECCO '21})}. \bibinfo{publisher}{Association for Computing Machinery}, \bibinfo{address}{New York, NY, USA}, \bibinfo{pages}{609–617}.
\newblock
\showISBNx{9781450383509}
\href{https://doi.org/10.1145/3449639.3459333}{doi:\nolinkurl{10.1145/3449639.3459333}}


\bibitem[Przewozniczek et~al\mbox{.}(2022)]%
        {dgga}
\bibfield{author}{\bibinfo{person}{Michal~W. Przewozniczek}, \bibinfo{person}{Renato Tin\'{o}s}, \bibinfo{person}{Bartosz Frej}, {and} \bibinfo{person}{Marcin~M. Komarnicki}.} \bibinfo{year}{2022}\natexlab{}.
\newblock \showarticletitle{On Turning Black - into Dark Gray-Optimization with the Direct Empirical Linkage Discovery and Partition Crossover}. In \bibinfo{booktitle}{\emph{Proceedings of the Genetic and Evolutionary Computation Conference}} (Boston, Massachusetts) \emph{(\bibinfo{series}{GECCO '22})}. \bibinfo{publisher}{Association for Computing Machinery}, \bibinfo{address}{New York, NY, USA}, \bibinfo{pages}{269–277}.
\newblock
\showISBNx{9781450392372}
\href{https://doi.org/10.1145/3512290.3528734}{doi:\nolinkurl{10.1145/3512290.3528734}}


\bibitem[Przewozniczek et~al\mbox{.}(2023)]%
        {FIHCwLL}
\bibfield{author}{\bibinfo{person}{Michal~Witold Przewozniczek}, \bibinfo{person}{Renato Tin\'{o}s}, {and} \bibinfo{person}{Marcin~Michal Komarnicki}.} \bibinfo{year}{2023}\natexlab{}.
\newblock \showarticletitle{First Improvement Hill Climber with Linkage Learning -- on Introducing Dark Gray-Box Optimization into Statistical Linkage Learning Genetic Algorithms}. In \bibinfo{booktitle}{\emph{Proceedings of the Genetic and Evolutionary Computation Conference}} (Lisbon, Portugal) \emph{(\bibinfo{series}{GECCO '23})}. \bibinfo{publisher}{ACM}, \bibinfo{pages}{946–954}.
\newblock
\showISBNx{9798400701191}


\bibitem[Przewozniczek et~al\mbox{.}(2017)]%
        {MuPPetsBaldwinEon}
\bibfield{author}{\bibinfo{person}{Michal~Witold Przewozniczek}, \bibinfo{person}{Krzysztof Walkowiak}, {and} \bibinfo{person}{Michal Aibin}.} \bibinfo{year}{2017}\natexlab{}.
\newblock \showarticletitle{The Evolutionary Cost of Baldwin Effect in the Routing and Spectrum Allocation Problem in Elastic Optical Networks}.
\newblock \bibinfo{journal}{\emph{Appl. Soft Comput.}} \bibinfo{volume}{52}, \bibinfo{number}{C} (\bibinfo{date}{March} \bibinfo{year}{2017}), \bibinfo{pages}{843--862}.
\newblock


\bibitem[Rahman and Szabó(2021)]%
        {envirRahman}
\bibfield{author}{\bibinfo{person}{Md Rahman} {and} \bibinfo{person}{György Szabó}.} \bibinfo{year}{2021}\natexlab{}.
\newblock \showarticletitle{Multi-objective Urban Land Use Optimization using Spatial Data: A systematic Review}.
\newblock \bibinfo{journal}{\emph{Sustainable Cities and Society}}  \bibinfo{volume}{74} (\bibinfo{date}{07} \bibinfo{year}{2021}), \bibinfo{pages}{103214}.
\newblock
\href{https://doi.org/10.1016/j.scs.2021.103214}{doi:\nolinkurl{10.1016/j.scs.2021.103214}}


\bibitem[Rodríguez and Coello(2021)]%
        {matingRestr}
\bibfield{author}{\bibinfo{person}{Amín V.~Bernabé Rodríguez} {and} \bibinfo{person}{Carlos A.~Coello Coello}.} \bibinfo{year}{2021}\natexlab{}.
\newblock \showarticletitle{An Ensemble of S-energy Based Mating Restrictions for Multi-Objective Evolutionary Algorithms}. In \bibinfo{booktitle}{\emph{2021 IEEE Symposium Series on Computational Intelligence (SSCI)}}. \bibinfo{pages}{1--8}.
\newblock
\href{https://doi.org/10.1109/SSCI50451.2021.9660112}{doi:\nolinkurl{10.1109/SSCI50451.2021.9660112}}


\bibitem[Rong et~al\mbox{.}(2020)]%
        {moeadStrongIGD}
\bibfield{author}{\bibinfo{person}{Miao Rong}, \bibinfo{person}{Dunwei Gong}, \bibinfo{person}{Witold Pedrycz}, {and} \bibinfo{person}{Ling Wang}.} \bibinfo{year}{2020}\natexlab{}.
\newblock \showarticletitle{A Multimodel Prediction Method for Dynamic Multiobjective Evolutionary Optimization}.
\newblock \bibinfo{journal}{\emph{IEEE Transactions on Evolutionary Computation}} \bibinfo{volume}{24}, \bibinfo{number}{2} (\bibinfo{year}{2020}), \bibinfo{pages}{290--304}.
\newblock
\href{https://doi.org/10.1109/TEVC.2019.2925358}{doi:\nolinkurl{10.1109/TEVC.2019.2925358}}


\bibitem[Schwaab et~al\mbox{.}(2017a)]%
        {envirSchwaab}
\bibfield{author}{\bibinfo{person}{Jonas Schwaab}, \bibinfo{person}{Kalyan Deb}, \bibinfo{person}{Erik Goodman}, \bibinfo{person}{Sven Lautenbach}, \bibinfo{person}{Maarten van Strien}, {and} \bibinfo{person}{Adrienne Grêt-Regamey}.} \bibinfo{year}{2017}\natexlab{a}.
\newblock \showarticletitle{Improving the performance of genetic algorithms for land-use allocation problems}.
\newblock \bibinfo{journal}{\emph{International Journal of Geographical Information Science}}  \bibinfo{volume}{32} (\bibinfo{date}{12} \bibinfo{year}{2017}), \bibinfo{pages}{1--24}.
\newblock
\href{https://doi.org/10.1080/13658816.2017.1419249}{doi:\nolinkurl{10.1080/13658816.2017.1419249}}


\bibitem[Schwaab et~al\mbox{.}(2017b)]%
        {problemTelSchwaab}
\bibfield{author}{\bibinfo{person}{Jonas Schwaab}, \bibinfo{person}{Kalyanmoy Deb}, \bibinfo{person}{Erik Goodman}, \bibinfo{person}{Sven Lautenbach}, \bibinfo{person}{Maarten {van Strien}}, {and} \bibinfo{person}{Adrienne Grêt-Regamey}.} \bibinfo{year}{2017}\natexlab{b}.
\newblock \showarticletitle{Reducing the loss of agricultural productivity due to compact urban development in municipalities of Switzerland}.
\newblock \bibinfo{journal}{\emph{Computers, Environment and Urban Systems}}  \bibinfo{volume}{65} (\bibinfo{year}{2017}), \bibinfo{pages}{162--177}.
\newblock
\showISSN{0198-9715}
\href{https://doi.org/10.1016/j.compenvurbsys.2017.06.005}{doi:\nolinkurl{10.1016/j.compenvurbsys.2017.06.005}}


\bibitem[Schwaab et~al\mbox{.}(2017c)]%
        {schwaab2017reducing}
\bibfield{author}{\bibinfo{person}{Jonas Schwaab}, \bibinfo{person}{Kalyanmoy Deb}, \bibinfo{person}{Erik Goodman}, \bibinfo{person}{Sven Lautenbach}, \bibinfo{person}{Maarten {van Strien}}, {and} \bibinfo{person}{Adrienne Grêt-Regamey}.} \bibinfo{year}{2017}\natexlab{c}.
\newblock \showarticletitle{Reducing the loss of agricultural productivity due to compact urban development in municipalities of Switzerland}.
\newblock \bibinfo{journal}{\emph{Computers, Environment and Urban Systems}}  \bibinfo{volume}{65} (\bibinfo{year}{2017}), \bibinfo{pages}{162--177}.
\newblock
\showISSN{0198-9715}
\href{https://doi.org/10.1016/j.compenvurbsys.2017.06.005}{doi:\nolinkurl{10.1016/j.compenvurbsys.2017.06.005}}


\bibitem[Schwaab et~al\mbox{.}(2018)]%
        {luoCoreSchwaab}
\bibfield{author}{\bibinfo{person}{Jonas Schwaab}, \bibinfo{person}{Kalyanmoy Deb}, \bibinfo{person}{Erik Goodman}, \bibinfo{person}{Sven Lautenbach}, \bibinfo{person}{Maarten~J. van Strien}, {and} \bibinfo{person}{Adrienne Grêt-Regamey}.} \bibinfo{year}{2018}\natexlab{}.
\newblock \showarticletitle{Improving the performance of genetic algorithms for land-use allocation problems}.
\newblock \bibinfo{journal}{\emph{International Journal of Geographical Information Science}} \bibinfo{volume}{32}, \bibinfo{number}{5} (\bibinfo{year}{2018}), \bibinfo{pages}{907--930}.
\newblock
\href{https://doi.org/10.1080/13658816.2017.1419249}{doi:\nolinkurl{10.1080/13658816.2017.1419249}}


\bibitem[Stewart and Janssen(2014)]%
        {envirStewart}
\bibfield{author}{\bibinfo{person}{Theodor Stewart} {and} \bibinfo{person}{Ron Janssen}.} \bibinfo{year}{2014}\natexlab{}.
\newblock \showarticletitle{A multiobjective GIS-based land use planning algorithm}.
\newblock \bibinfo{journal}{\emph{Computers, Environment and Urban Systems}}  \bibinfo{volume}{46} (\bibinfo{date}{07} \bibinfo{year}{2014}).
\newblock
\href{https://doi.org/10.1016/j.compenvurbsys.2014.04.002}{doi:\nolinkurl{10.1016/j.compenvurbsys.2014.04.002}}


\bibitem[Strassburg et~al\mbox{.}(2020)]%
        {envirStrass}
\bibfield{author}{\bibinfo{person}{{Bernardo B.N.} Strassburg}, \bibinfo{person}{Alvaro Iribarrem}, \bibinfo{person}{{Hawthorne L.} Beyer}, \bibinfo{person}{{Carlos Leandro} Cordeiro}, \bibinfo{person}{Renato Crouzeilles}, \bibinfo{person}{{Catarina C.} Jakovac}, \bibinfo{person}{Andr{\'e} {Braga Junqueira}}, \bibinfo{person}{Eduardo Lacerda}, \bibinfo{person}{{Agnieszka E.} Latawiec}, \bibinfo{person}{Andrew Balmford}, \bibinfo{person}{{Thomas M.} Brooks}, \bibinfo{person}{{Stuart H.M.} Butchart}, \bibinfo{person}{{Robin L.} Chazdon}, \bibinfo{person}{{Karl Heinz} Erb}, \bibinfo{person}{Pedro Brancalion}, \bibinfo{person}{Graeme Buchanan}, \bibinfo{person}{David Cooper}, \bibinfo{person}{Sandra D{\'i}az}, \bibinfo{person}{{Paul F.} Donald}, \bibinfo{person}{Valerie Kapos}, \bibinfo{person}{David Lecl{\`e}re}, \bibinfo{person}{Lera Miles}, \bibinfo{person}{Michael Obersteiner}, \bibinfo{person}{Christoph Plutzar}, \bibinfo{person}{{Carlos Alberto} {de M. Scaramuzza}}, \bibinfo{person}{{Fabio R.} Scarano},
  {and} \bibinfo{person}{Piero Visconti}.} \bibinfo{year}{2020}\natexlab{}.
\newblock \showarticletitle{Global priority areas for ecosystem restoration}.
\newblock \bibinfo{journal}{\emph{Nature}} \bibinfo{volume}{586}, \bibinfo{number}{7831} (\bibinfo{date}{29 Oct.} \bibinfo{year}{2020}), \bibinfo{pages}{724--729}.
\newblock
\showISSN{0028-0836}
\href{https://doi.org/10.1038/s41586-020-2784-9}{doi:\nolinkurl{10.1038/s41586-020-2784-9}}


\bibitem[Thierens and Bosman(2013)]%
        {ltga}
\bibfield{author}{\bibinfo{person}{Dirk Thierens} {and} \bibinfo{person}{Peter~A.N. Bosman}.} \bibinfo{year}{2013}\natexlab{}.
\newblock \showarticletitle{Hierarchical Problem Solving with the Linkage Tree Genetic Algorithm}. In \bibinfo{booktitle}{\emph{Proceedings of the 15th Annual Conference on Genetic and Evolutionary Computation}} \emph{(\bibinfo{series}{GECCO '13})}. \bibinfo{publisher}{ACM}, \bibinfo{pages}{877--884}.
\newblock


\bibitem[Tin\'{o}s et~al\mbox{.}(2023)]%
        {wVIG}
\bibfield{author}{\bibinfo{person}{Renato Tin\'{o}s}, \bibinfo{person}{Michal Przewozniczek}, \bibinfo{person}{Darrell Whitley}, {and} \bibinfo{person}{Francisco Chicano}.} \bibinfo{year}{2023}\natexlab{}.
\newblock \showarticletitle{Genetic Algorithm with Linkage Learning}. In \bibinfo{booktitle}{\emph{Proceedings of the Genetic and Evolutionary Computation Conference}} (Lisbon, Portugal) \emph{(\bibinfo{series}{GECCO '23})}. \bibinfo{publisher}{Association for Computing Machinery}, \bibinfo{address}{New York, NY, USA}, \bibinfo{pages}{981–989}.
\newblock
\showISBNx{9798400701191}
\href{https://doi.org/10.1145/3583131.3590349}{doi:\nolinkurl{10.1145/3583131.3590349}}


\bibitem[Tin\'{o}s et~al\mbox{.}(2022)]%
        {ilsDLED}
\bibfield{author}{\bibinfo{person}{Renato Tin\'{o}s}, \bibinfo{person}{Michal~W. Przewozniczek}, {and} \bibinfo{person}{Darrell Whitley}.} \bibinfo{year}{2022}\natexlab{}.
\newblock \showarticletitle{Iterated Local Search with Perturbation Based on Variables Interaction for Pseudo-Boolean Optimization}. In \bibinfo{booktitle}{\emph{Proceedings of the Genetic and Evolutionary Computation Conference}} (Boston, Massachusetts) \emph{(\bibinfo{series}{GECCO '22})}. \bibinfo{publisher}{ACM}, \bibinfo{pages}{296–304}.
\newblock
\showISBNx{9781450392372}


\bibitem[Tin\'{o}s et~al\mbox{.}(2015)]%
        {pxForBinary}
\bibfield{author}{\bibinfo{person}{Renato Tin\'{o}s}, \bibinfo{person}{Darrell Whitley}, {and} \bibinfo{person}{Francisco Chicano}.} \bibinfo{year}{2015}\natexlab{}.
\newblock \showarticletitle{Partition Crossover for Pseudo-Boolean Optimization}. In \bibinfo{booktitle}{\emph{Proceedings of the 2015 ACM Conference on Foundations of Genetic Algorithms XIII}} (Aberystwyth, United Kingdom) \emph{(\bibinfo{series}{FOGA '15})}. \bibinfo{publisher}{Association for Computing Machinery}, \bibinfo{address}{New York, NY, USA}, \bibinfo{pages}{137–149}.
\newblock
\showISBNx{9781450334341}
\href{https://doi.org/10.1145/2725494.2725497}{doi:\nolinkurl{10.1145/2725494.2725497}}


\bibitem[Tomczyk and Kadziński(2020)]%
        {moeadStrong}
\bibfield{author}{\bibinfo{person}{Michał~K. Tomczyk} {and} \bibinfo{person}{Miłosz Kadziński}.} \bibinfo{year}{2020}\natexlab{}.
\newblock \showarticletitle{Decomposition-Based Interactive Evolutionary Algorithm for Multiple Objective Optimization}.
\newblock \bibinfo{journal}{\emph{IEEE Transactions on Evolutionary Computation}} \bibinfo{volume}{24}, \bibinfo{number}{2} (\bibinfo{year}{2020}), \bibinfo{pages}{320--334}.
\newblock
\href{https://doi.org/10.1109/TEVC.2019.2915767}{doi:\nolinkurl{10.1109/TEVC.2019.2915767}}


\bibitem[Verel et~al\mbox{.}(2018)]%
        {walshSurBin}
\bibfield{author}{\bibinfo{person}{S{\'e}bastien Verel}, \bibinfo{person}{Bilel Derbel}, \bibinfo{person}{Arnaud Liefooghe}, \bibinfo{person}{Hern{\'a}n Aguirre}, {and} \bibinfo{person}{Kiyoshi Tanaka}.} \bibinfo{year}{2018}\natexlab{}.
\newblock \showarticletitle{A Surrogate Model Based on Walsh Decomposition for Pseudo-Boolean Functions}. In \bibinfo{booktitle}{\emph{Parallel Problem Solving from Nature -- PPSN XV}}, \bibfield{editor}{\bibinfo{person}{Anne Auger}, \bibinfo{person}{Carlos~M. Fonseca}, \bibinfo{person}{Nuno Louren{\c{c}}o}, \bibinfo{person}{Penousal Machado}, \bibinfo{person}{Lu{\'i}s Paquete}, {and} \bibinfo{person}{Darrell Whitley}} (Eds.). \bibinfo{publisher}{Springer International Publishing}, \bibinfo{address}{Cham}, \bibinfo{pages}{181--193}.
\newblock
\showISBNx{978-3-319-99259-4}


\bibitem[Whitley(2019)]%
        {whitleyNext}
\bibfield{author}{\bibinfo{person}{D. Whitley}.} \bibinfo{year}{2019}\natexlab{}.
\newblock \showarticletitle{Next generation genetic algorithms: a user’s guide and tutorial}.
\newblock In \bibinfo{booktitle}{\emph{Handbook of Metaheuristics}}. \bibinfo{publisher}{Springer}, \bibinfo{pages}{245--274}.
\newblock


\bibitem[Whitley et~al\mbox{.}(2016)]%
        {GrayBoxWhitley}
\bibfield{author}{\bibinfo{person}{L.~Darrell Whitley}, \bibinfo{person}{Francisco Chicano}, {and} \bibinfo{person}{Brian~W. Goldman}.} \bibinfo{year}{2016}\natexlab{}.
\newblock \showarticletitle{Gray Box Optimization for Mk Landscapes Nk Landscapes and Max-Ksat}.
\newblock \bibinfo{journal}{\emph{Evol. Comput.}} \bibinfo{volume}{24}, \bibinfo{number}{3} (\bibinfo{date}{Sept.} \bibinfo{year}{2016}), \bibinfo{pages}{491–519}.
\newblock
\showISSN{1063-6560}
\href{https://doi.org/10.1162/EVCO_a_00184}{doi:\nolinkurl{10.1162/EVCO_a_00184}}


\bibitem[Yu and Goldberg(2004)]%
        {missingIsBad}
\bibfield{author}{\bibinfo{person}{Tian-Li Yu} {and} \bibinfo{person}{David~E. Goldberg}.} \bibinfo{year}{2004}\natexlab{}.
\newblock \showarticletitle{Toward an Understanding of the Quality and Efficiency of Model Building for Genetic Algorithms}. In \bibinfo{booktitle}{\emph{Genetic and Evolutionary Computation -- GECCO 2004}}, \bibfield{editor}{\bibinfo{person}{Kalyanmoy Deb}} (Ed.). \bibinfo{publisher}{Springer Berlin Heidelberg}, \bibinfo{address}{Berlin, Heidelberg}, \bibinfo{pages}{367--378}.
\newblock
\showISBNx{978-3-540-24855-2}


\bibitem[{Zhang} and {Li}(2007)]%
        {moead}
\bibfield{author}{\bibinfo{person}{Q. {Zhang}} {and} \bibinfo{person}{H. {Li}}.} \bibinfo{year}{2007}\natexlab{}.
\newblock \showarticletitle{MOEA/D: A Multiobjective Evolutionary Algorithm Based on Decomposition}.
\newblock \bibinfo{journal}{\emph{IEEE Transactions on Evolutionary Computation}} \bibinfo{volume}{11}, \bibinfo{number}{6} (\bibinfo{date}{Dec} \bibinfo{year}{2007}), \bibinfo{pages}{712--731}.
\newblock
\showISSN{1941-0026}
\href{https://doi.org/10.1109/TEVC.2007.892759}{doi:\nolinkurl{10.1109/TEVC.2007.892759}}


\end{thebibliography}
